\documentclass{article}%
\usepackage{amsmath}
\usepackage{amsfonts}
\usepackage{amssymb}
\usepackage{graphicx}
\usepackage{algorithmic}
\usepackage{algorithm}%
\providecommand{\U}[1]{\protect\rule{.1in}{.1in}}

\newtheorem{remark}{Remark}

\begin{document}

\title{LARF: Two-level Attention-based Random Forests with a Mixture of Contamination Models}
\author{Andrei V. Konstantinov and Lev V. Utkin\\Peter the Great St.Petersburg Polytechnic University\\St.Petersburg, Russia\\e-mail: andrue.konst@gmail.com, lev.utkin@gmail.com}
\date{}
\maketitle

\begin{abstract}
New models of the attention-based random forests called LARF (Leaf
Attention-based Random Forest) are proposed. The first idea behind the models
is to introduce a two-level attention, where one of the levels is the
\textquotedblleft leaf\textquotedblright\ attention and the attention
mechanism is applied to every leaf of trees. The second level is the tree
attention depending on the \textquotedblleft leaf\textquotedblright%
\ attention. The second idea is to replace the softmax operation in the
attention with the weighted sum of the softmax operations with different
parameters. It is implemented by applying a mixture of the Huber's
contamination models and can be regarded as an analog of the multi-head
attention with \textquotedblleft heads\textquotedblright\ defined by selecting
a value of the softmax parameter. Attention parameters are simply trained by
solving the quadratic optimization problem. To simplify the tuning process of
the models, it is proposed to make the tuning contamination parameters to be
training and to compute them by solving the quadratic optimization problem.
Many numerical experiments with real datasets are performed for studying
LARFs. The code of proposed algorithms can be found in https://github.com/andruekonst/leaf-attention-forest.


\end{abstract}

\section{Introduction}

Several crucial improvements of neural networks have been made in recent
years. One of them is the attention mechanism which has played an important
role in many machine learning areas, including the natural language processing
models, the computer vision, etc.
\cite{Chaudhari-etal-2019,Correia-Colombini-21a,Correia-Colombini-21,Lin-Wang-etal-21,Niu-Zhong-Yu-21}%
. The idea behind the attention mechanism is to assign weights to features or
examples in accordance with their importance and their impact on the model
predictions. At the same time, the success of the attention models as
components of neural network motivates to extend this approach to other
machine learning models different from neural networks, for example, to random
forests (RFs) \cite{Breiman-2001}. Following this idea, a new model called the
attention-based random forest (ABRF), which incorporates the attention
mechanism into ensemble-based models such as RFs and the gradient boosting
machine \cite{Friedman-2001,Friedman-2002} has been developed
\cite{Konstantinov-Utkin-22d,Utkin-Konstantinov-22}. The ABRF models stems
from the interesting interpretation \cite{Chaudhari-etal-2019,Zhang2021dive}
of the attention mechanism through the Nadaraya-Watson kernel regression model
\cite{Nadaraya-1964,Watson-1964}. According to
\cite{Konstantinov-Utkin-22d,Utkin-Konstantinov-22}, attention weights in the
Nadaraya-Watson regression are assigned to decision trees in a RF depending on
examples which fall into leaves of trees. Weights in ABRF have trainable
parameters and use the Huber's $\epsilon$-contamination model \cite{Huber81}
for defining the attention weights such that each attention weight consists of
two parts: the softmax operation with the tuning coefficient $1-\epsilon$ and
the trainable bias of the softmax weight with coefficient $\epsilon$. One of
the improvements of ABRF, which has been proposed in
\cite{Utkin-Konstantinov-22c}, is based on joint incorporating self-attention
and attention mechanisms into the RF. The proposed models outperform ABRF, but
this outperformance is not sufficient. The model for several dataset provided
inferior results. Therefore, we propose a set of models which can be regarded
as extensions of ABRF and are based on two main ideas.

The first idea is to introduce a two-level attention, where one of the levels
is the \textquotedblleft leaf\textquotedblright\ attention, i.e., the
attention mechanism is applied to every leaf of trees. As a result, we get the
attention weights assigned to leaves and the attention weights assigned to
trees. At that, the attention weights of trees depend on the corresponding
weights of leaves belonging to these trees. In other words, the attention at
the second level depends on the attention at the first level, i.e., we get the
attention of the attention. Due to the \textquotedblleft
leaf\textquotedblright\ attention, the proposed model will be abbreviated as
LARF (Leaf Attention-based Random Forest).

One of the peculiarities of LARFs is the use of a mixture of the Huber's
$\epsilon$-contamination models instead of the single contamination model as
it has been done in ABRF. This peculiarity stems from the second idea behind
the model to take into account the softmax operation with different parameters
simultaneously. In fact, we replace the standard softmax operation by the
weighted sum of the softmax operations with different parameters. With this
idea, we achieve two goals. First of all, we partially solve the problem of
tuning parameters of the softmax operations which are a part of attention
operations. Each value of the tuning parameter from a predefined set (from a
predefined grid) is used in a separate softmax operation. Then weights of the
softmax operations in the sum are trained jointly with training other
parameters. This approach can also be interpreted as the linear approximation
of the softmax operations with trainable weights and with different values of
tuning parameters. However, a more interesting goal is that some analog of the
multi-head attention \cite{Vaswani-etal-17} is implemented by using the
mixture of contamination models where \textquotedblleft
heads\textquotedblright\ are defined by selecting a value of the corresponding
softmax operation parameter.

Additionally, in contrast to ABRF \cite{Utkin-Konstantinov-22} where the
contamination parameter $\epsilon$ of the Huber's model was a tuning
parameter, the LARF model considers this parameter as the training one. This
allows us to significantly reduce the model tuning time and to avoid
enumeration of the parameter values in accordance with a grid. The same is
implemented for the mixture of the Huber's models.

Different configurations of LARF produce a set of models which depend on
trainable parameters of the two-level attention and its implementation.

We investigate two types of RFs in experiments: original RFs and Extremely
Randomized Trees (ERT) \cite{Geurts-etal-06}. According to
\cite{Geurts-etal-06}, the ERT algorithm chooses a split point randomly for
each feature at each node and then selects the best split among these.

Our contributions can be summarized as follows:

\begin{enumerate}
\item New two-level attention-based RF models are proposed, where the
attention mechanism at the first level is applied to every leaf of trees, the
attention at the second level incorporates the \textquotedblleft
leaf\textquotedblright\ attention and is applied to trees. Training of the
two-level attention is reduced to solving the standard quadratic optimization problem.

\item A mixture of the Huber's $\epsilon$-contamination models is used to
implement the attention mechanism at the second level. The mixture allows us
to replace a set of tuning attention parameters (the temperature parameters of
the softmax operations) with trainable parameters whose optimal values are
computed by solving the quadratic optimization problem. Moreover, this
approach can be regarded as an analog of the multi-head attention.

\item An approach is proposed to make the tuning contamination parameters
($\epsilon$ parameters) in the mixture of the $\epsilon$-contamination models
to be training. Their optimal values are also computed by solving the
quadratic optimization problem.

\item Many numerical experiments with real datasets are performed for studying
LARFs. They demonstrate outperforming results of some modifications of LARF.
The code of proposed algorithms can be found in https://github.com/andruekonst/leaf-attention-forest.
\end{enumerate}

The paper is organized as follows. Related work can be found in Section 2. A
brief introduction to the attention mechanism as the Nadaraya-Watson kernel
regression is given in Section 3. A general approach to incorporating the
two-level attention mechanism into the RF is provided in Section 4. Ways for
implementation of the two-level attention mechanism and constructing several
attention-based models by using the mixture of the Huber's $\epsilon
$-contamination models are considered in Section 5. Numerical experiments with
real data illustrating properties of the proposed models are provided in
Section 6. Concluding remarks can be found in Section 7.

\section{Related work}

\textbf{Attention mechanism}. Due to the great efficiency of machine learning
models with the attention mechanisms, interest in the different
attention-based models has increased significantly in recent years. As a
result, many attention models have been proposed to improve the performance of
machine learning algorithms. The most comprehensive analysis and description
of various attention-based models can be found in interesting surveys
\cite{Chaudhari-etal-2019,Correia-Colombini-21a,Correia-Colombini-21,Lin-Wang-etal-21,Niu-Zhong-Yu-21,Liu-Huang-etal-21}%
.

It is important to note that parametric attention models as parts of neural
networks are mainly trained by applying the gradient-based algorithms which
lead to computational problems when training is carried out through the
softmax function. Many approaches have been proposed to cope with this
problem. A large part of approaches is based on some kinds of linear
approximation of the softmax attention of
\cite{Choromanski-etal-21,Choromanski-etal-21a,Ma-Kong-etal-21,Schlag-etal-2021}%
. Another part of the approaches is based on random feature methods to
approximate the softmax function \cite{Liu-Huang-etal-21,Peng-Pappas-etal-21}.

Another improvement of the attention-based models is to use the self-attention
which was proposed in \cite{Vaswani-etal-17} as a crucial component of neural
networks called Transformers. The self-attention models have been also studied
in surveys
\cite{Lin-Wang-etal-21,Brauwers-Frasincar-22,Goncalves-etal-2022,Santana-Colombini-21,Soydaner-22,Xu-Wei-etal-22}%
. This is only a small part of all works devoted to the attention and
self-attention mechanisms.

It should be noted that the aforementioned models are implemented as neural
networks, and they have not been studied for application to other machine
learning models, for example, to RFs. Attempts to incorporate the attention
and self-attention mechanisms into the RF and the gradient boosting machine
were made in
\cite{Konstantinov-Utkin-22d,Utkin-Konstantinov-22,Utkin-Konstantinov-22c}.
Following these works, we extend the proposed models to improve the
attention-based models. Moreover, we propose the attention models which do not
use the gradient-based algorithms for computing optimal attention parameters.
The training process of the models is based on solving standard quadratic
optimization problems.

\textbf{Weighted RFs}. Many approaches have been proposed in recent years to
improve RFs. One of the important approaches is based on assignment of weights
to decision trees in the RF. This approach is implemented in various
algorithms
\cite{Kim-Kim-Moon-Ahn-2011,Ronao-Cho-2015,Utkin-Konstantinov-etal-20,Utkin-etal-2019,Winham-etal-2013,Xuan-etal-18,Zhang-Wang-21}%
. However, most these algorithms have an disadvantage. The weights are
assigned to trees independently of examples, i.e., each weight characterizes
trees on the average over all training examples and does not take into account
any feature vector. Moreover, the weights do not have training parameters
which usually make the model more flexible and accurate.

\textbf{Contamination model in attention mechanisms}. There are several models
which use imprecise probabilities in order to model the lack of sufficient
training data. One of the first models is the so-called Credal Decision Tree,
which is based on applying the imprecise probability theory to classification
and proposed in \cite{Abellan-Moral-2003}. Following this work, a number of
models based on imprecise probabilities were presented in
\cite{Abellan-etal-2017,Abellan-etal-2018,Mantas-Abellan-2014,Moral-Garcia-etal-2020}
where the imprecise Dirichlet model is used. This model can be regarded as a
reparametrization of the imprecise $\epsilon$-contamination model which is
applied to LARF. The imprecise $\epsilon$-contamination model has been also
applied to machine learning methods, for example, to the support vector
machine \cite{Utkin-Wiencierz-13} or to the RF
\cite{Utkin-Kovalev-Coolen-2020}. The attention-based RF applying the
imprecise $\epsilon$-contamination model to the parametric attention mechanism
was proposed in \cite{Utkin-Konstantinov-22,Utkin-Konstantinov-22c}. However,
there were no other works which use the imprecise models in order to implement
the attention mechanism.

\section{Nadaraya-Watson regression and the attention mechanism}

A basis of the attention mechanism can be considered in the framework of the
Nadaraya-Watson kernel regression model \cite{Nadaraya-1964,Watson-1964} which
estimates a function $f$ as a locally weighted average using a kernel as a
weighting function. Suppose that the dataset is represented by $n$ examples
$(\mathbf{x}_{1},y_{1}),...,(\mathbf{x}_{n},y_{n})$, where $\mathbf{x}%
_{i}=(x_{i1},...,x_{im})\in\mathbb{R}^{m}$ is a feature vector consisting of
$m$ features; $y_{i}\in\mathbb{R}$ is a regression output. The regression task
is to construct a regressor $f:\mathbb{R}^{m}\rightarrow\mathbb{R}$ which can
predict the output value $\tilde{y}$ of a new observation $\mathbf{x}$, using
the dataset.

The Nadaraya-Watson kernel regression estimates the regression output
$\tilde{y}$ corresponding to a new input feature vector $\mathbf{x}$ as
follows \cite{Nadaraya-1964,Watson-1964}:%
\begin{equation}
\tilde{y}=\sum_{i=1}^{n}\alpha(\mathbf{x},\mathbf{x}_{i})y_{i},
\end{equation}
where weight $\alpha(\mathbf{x},\mathbf{x}_{i})$ conforms with relevance of
the feature vector $\mathbf{x}_{i}$ to the vector $\mathbf{x}$.

It can be seen from the above that the Nadaraya-Watson regression model
estimates $\tilde{y}$ as a weighted sum of training outputs $y_{i}$ from the
dataset such that their weights depend on location of $\mathbf{x}_{i}$
relative to $\mathbf{x}$. This mean that the closer $\mathbf{x}_{i}$ to
$\mathbf{x}$, the greater the weight assigned to $y_{i}$.

According to the Nadaraya-Watson kernel regression
\cite{Nadaraya-1964,Watson-1964}, weights can be defined by means of a kernel
$K$ as a function of the distance between vectors $\mathbf{x}_{i}$ and
$\mathbf{x}$. The kernel estimates how $\mathbf{x}_{i}$ is close to
$\mathbf{x}$. Then the weight is written as follows:
\begin{equation}
\alpha(\mathbf{x},\mathbf{x}_{i})=\frac{K(\mathbf{x},\mathbf{x}_{i})}%
{\sum_{j=1}^{n}K(\mathbf{x},\mathbf{x}_{j})}.
\end{equation}

One of the popular kernel is the Gaussian kernel. It produces weights of the
form:
\begin{equation}
\alpha(\mathbf{x},\mathbf{x}_{i})=\text{$\sigma$}\left(  -\frac{\left\Vert
\mathbf{x}-\mathbf{x}_{i}\right\Vert ^{2}}{\tau}\right)  ,
\end{equation}
where $\tau$ is a tuning (temperature) parameter; $\sigma\left(  \cdot\right)
$ is a notation of the softmax operation.

In terms of the attention mechanism \cite{Bahdanau-etal-14}, vector
$\mathbf{x}$, vectors $\mathbf{x}_{i}$, outputs $y_{i}$ and weight
$\alpha(\mathbf{x},\mathbf{x}_{i})$ are called as the \textit{query},
\textit{keys}, \textit{values} and the \textit{attention weight},
respectively. Weights $\alpha(\mathbf{x},\mathbf{x}_{i})$ can be extended by
incorporating trainable parameters. In particular, parameter $\tau$ can be
also regarded as the trainable parameter.

Many forms of parametric attention weights, which also define the attention
mechanisms, have been proposed, for example, the additive attention
\cite{Bahdanau-etal-14}, the multiplicative or dot-product attention
\cite{Vaswani-etal-17,Luong-etal-2015}. We will consider also the attention
weights based on the Gaussian kernels, i.e., producing the softmax operation.
However, the parametric forms of the attention weights will be quite different
from many popular attention operations.

\section{Two-level attention-based random forest}

One of the powerful machine learning models handling with tabular data is the
RF which can be regarded as an ensemble of $T$ decision trees such that each
tree is trained on a subset of examples randomly selected from the training
set. In the original RF, the final RF prediction $\tilde{y}$ for a testing
example $\mathbf{x}$ is determined by averaging predictions $\tilde{y}%
_{1},...,\tilde{y}_{T}$ obtained for all trees.

Let $\mathcal{J}_{k}(\mathbf{x})$ be the index set of examples which fall into
the same leaf in the $k$-th tree as $\mathbf{x}$. One of the ways to construct
the attention-based RF is to introduce the mean vector $\mathbf{A}%
_{k}(\mathbf{x)}$ defined as the mean of training vectors $\mathbf{x}_{j}$
which fall into the same leaf as $\mathbf{x}$. However, this simple definition
can be extended by incorporating the Nadaraya-Watson regression into the leaf.
In this case, we can write
\begin{equation}
\mathbf{A}_{k}(\mathbf{x)}=\sum_{j\in\mathcal{J}_{k}(\mathbf{x})}\mu\left(
\mathbf{x},\mathbf{x}_{j}\right)  \mathbf{x}_{j}, \label{RF_Att_L_20}%
\end{equation}
where $\mu\left(  \mathbf{x},\mathbf{x}_{j}\right)  $ is the attention weight
in accordance with the Nadaraya-Watson kernel regression.

In fact, (\ref{RF_Att_L_20}) can be regarded as the self-attention. The idea
behind (\ref{RF_Att_L_20}) is that we find the mean value of $\mathbf{x}$ by
assigning weights to training examples which fall into the corresponding leaf
in accordance with their vicinity to the vector $\mathbf{x}$.

In the same way, we can define the mean value of regression outputs
corresponding to examples falling into the same leaf as $\mathbf{x}$:%
\begin{equation}
B_{k}(\mathbf{x)}=\sum_{j\in\mathcal{J}_{k}(\mathbf{x})}\mu\left(
\mathbf{x},\mathbf{x}_{j}\right)  y_{j}. \label{RF_Att_L_21}%
\end{equation}

Expression (\ref{RF_Att_L_21}) can be regarded as the attention. The idea
behind (\ref{RF_Att_L_21}) is to get the prediction provided by the
corresponding leaf by using the standard attention mechanism or the
Nadaraya-Watson regression. In other words, we weigh predictions provided by
the $k$-th leaf of a tree in accordance with the distance between the feature
vector $\mathbf{x}$, which falls into the $k$-th leaf, and all feature vectors
$\mathbf{x}_{j}$ which fall into the same leaf. It should be noted that the
original regression tree provides the averaged prediction, i.e., it
corresponds to the case when all $\mu\left(  \mathbf{x},\mathbf{x}_{j}\right)
$ are identical for all $j\in\mathcal{J}_{k}(\mathbf{x})$ and equal to
$1/\#\mathcal{J}_{k}(\mathbf{x})$.

We suppose that the attention mechanisms used above are non-parametric. This
implies that weights do not have trainable parameters. It is assumed that%

\begin{equation}
\sum_{j\in\mathcal{J}_{k}(\mathbf{x})}\mu\left(  \mathbf{x},\mathbf{x}%
_{j}\right)  =1. \label{RF_Att_L_22}%
\end{equation}

The \textquotedblleft leaf\textquotedblright\ attention introduced above can
be regarded as the first-level attention in a hierarchy of the attention
mechanisms. It characterizes how the feature vector $\mathbf{x}$ fits the
corresponding tree.

If we suppose that the whole RF consists of $T$ decision trees, then the set
of $\mathbf{A}_{k}(\mathbf{x)}$, $k=1,...,T$, in the framework of the
attention mechanism can be regarded as a set of keys for every $\mathbf{x}$,
the set of $B_{k}(\mathbf{x)}$, $k=1,...,T$, can be regarded as a set of
values. This implies that the final prediction $\tilde{y}$ of the RF can be
computed by using the Nadaraya-Watson regression, namely,
\begin{equation}
\tilde{y}=f(\mathbf{x},\mathbf{w})=\sum_{k=1}^{T}\alpha\left(  \mathbf{x}%
,\mathbf{A}_{k}(\mathbf{x}),\mathbf{w}\right)  B_{k}(\mathbf{x)}.
\label{RF_Att_L_47}%
\end{equation}

Here $\alpha\left(  \mathbf{x},\mathbf{A}_{k}(\mathbf{x)},\mathbf{w}\right)  $
is the attention weight with vector $\mathbf{w}=(w_{1},...,w_{T})$ of
trainable parameters belonging to a set $\mathcal{W}$ such that they are
assigned to each tree. The attention weight $\alpha$ is defined by the
distance between $\mathbf{x}$ and $\mathbf{A}_{k}(\mathbf{x)}$. It is assumed
due to properties of the attention weights in the Nadaraya-Watson regression
that there holds
\begin{equation}
\sum_{k=1}^{T}\alpha\left(  \mathbf{x},\mathbf{A}_{k}(\mathbf{x}%
),\mathbf{w}\right)  =1. \label{RF_Att_L_101}%
\end{equation}

The above \textquotedblleft random forest\textquotedblright\ attention can be
regarded as the second-level attention which assigned weights to trees in
accordance with their impact on the RF prediction corresponding to
$\mathbf{x}$.

The main idea behind the approach is to use the above attention mechanisms
jointly. After substituting (\ref{RF_Att_L_20}) and (\ref{RF_Att_L_21}) into
(\ref{RF_Att_L_47}), we get
\begin{equation}
\tilde{y}(\mathbf{x})=f(\mathbf{x},\mathbf{w})=\sum_{k=1}^{T}\alpha\left(
\mathbf{x},\mathbf{A}_{k}(\mathbf{x)},\mathbf{w}\right)  \sum_{j\in
\mathcal{J}_{k}(\mathbf{x})}y_{j}\mu\left(  \mathbf{x},\mathbf{x}_{j}\right)
, \label{RF_Att_123}%
\end{equation}
or
\begin{equation}
\tilde{y}(\mathbf{x})=\sum_{k=1}^{T}\alpha\left(  \mathbf{x},\sum
_{i\in\mathcal{J}_{k}(\mathbf{x})}\mu\left(  \mathbf{x},\mathbf{x}_{i}\right)
\mathbf{x}_{i},\mathbf{w}\right)  \sum_{j\in\mathcal{J}_{k}(\mathbf{x})}%
y_{j}\mu\left(  \mathbf{x},\mathbf{x}_{j}\right)  .
\end{equation}

\begin{figure}
[ptb]
\begin{center}
\includegraphics[
height=2.6463in,
width=5.662in
]%
{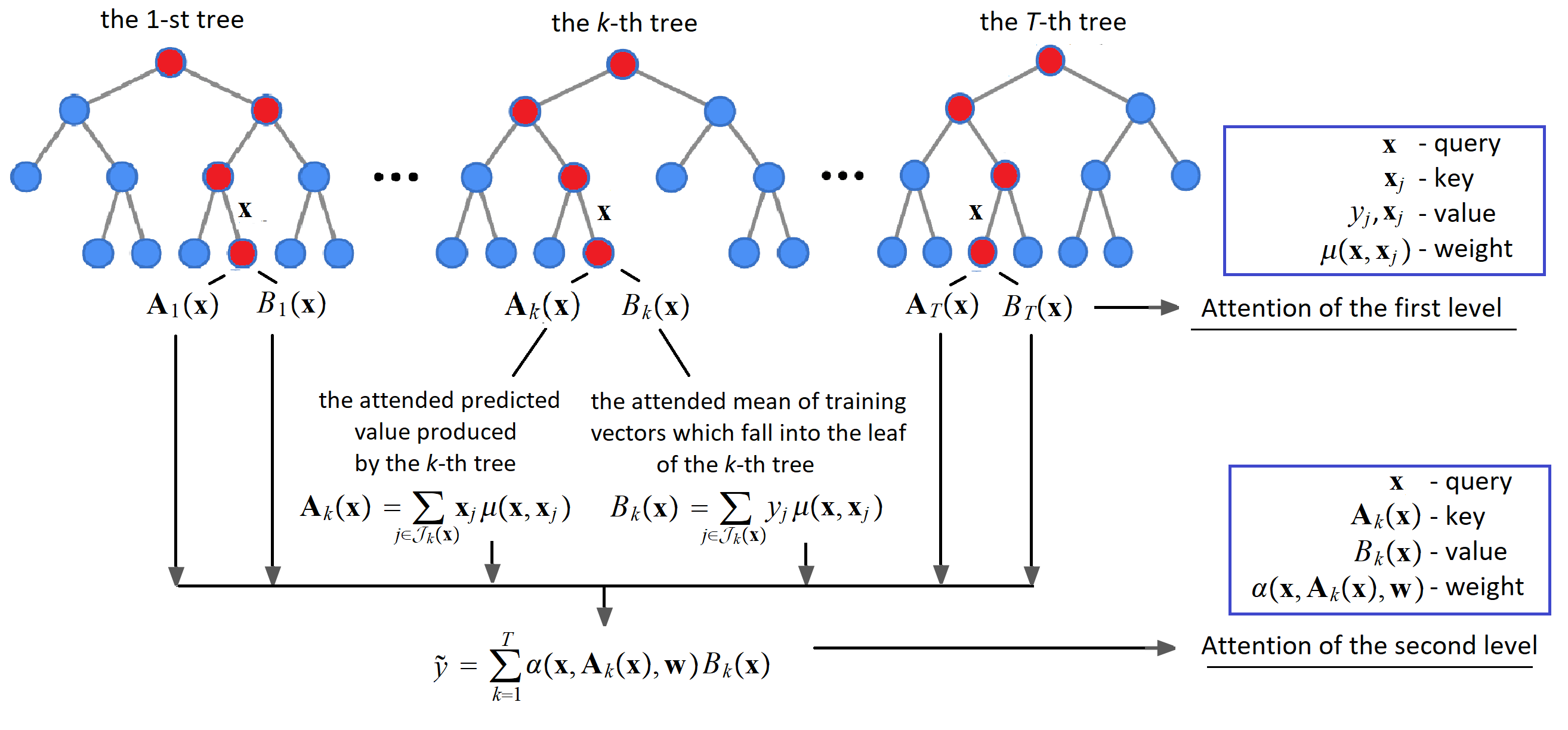}%
\caption{A scheme of the proposed two-level hierarchical attention model
applied to the RF}%
\label{f:nw_leaves}%
\end{center}
\end{figure}

A scheme of two-level attention is shown in Fig. \ref{f:nw_leaves}. It is
clearly seen from Fig. \ref{f:nw_leaves} how the attention at the second level
depends on the \textquotedblleft leaf\textquotedblright\ attention at the
first level.

In sum, we get the trainable attention-based RF with parameters $\mathbf{w}$,
which are defined by minimizing the expected loss function over set
$\mathcal{W}$ of parameters, respectively, as follows:
\begin{equation}
\mathbf{w}_{opt}=\arg\min_{\mathbf{w\in}\mathcal{W}}~\sum_{s=1}^{n}L\left(
\tilde{y}(\mathbf{x}_{s}),y_{s},\mathbf{w}\right)  . \label{RF_Att_49}%
\end{equation}

The loss function can be rewritten as
\begin{align}
&  \sum_{s=1}^{n}L\left(  \tilde{y}(\mathbf{x}_{s}),y_{s},\mathbf{w}\right)
=\sum_{s=1}^{n}\left(  y_{s}-\tilde{y}(\mathbf{x}_{s})\right)  ^{2}\nonumber\\
&  =\sum_{s=1}^{n}\left(  y_{s}-\sum_{k=1}^{T}\sum_{j\in\mathcal{J}%
_{k}(\mathbf{x}_{s})}y_{j}\mu\left(  \mathbf{x}_{s},\mathbf{x}_{j}\right)
\cdot\alpha\left(  \mathbf{x}_{s},\sum_{i\in\mathcal{J}_{k}(\mathbf{x}_{s}%
)}\mu\left(  \mathbf{x}_{s},\mathbf{x}_{i}\right)  \mathbf{x}_{i}%
,\mathbf{w}\right)  \right)  ^{2}. \label{RF_Att_50}%
\end{align}

Optimal trainable parameters $\mathbf{w}$ are computed depending on forms of
attention weights $\alpha$ in optimization problem (\ref{RF_Att_50}). It
should be noted that problem (\ref{RF_Att_50}) may be complex from the
computation point of view. Therefore, one of our results is to propose such a
form of attention weights $\alpha$ that makes problem (\ref{RF_Att_50}) a
convex quadratic optimization problem whose solution does not meet any difficulties.%

\begin{remark}
It is important to point out that the additional sets of trainable parameters
can be introduced into the definition of attention weights $\mu\left(
\mathbf{x},\mathbf{x}_{j}\right)  $. On the one hand, we get a more flexible
attention mechanisms in this case due to the parametrization of training
weights $\mu\left(  \mathbf{x},\mathbf{x}_{j}\right)  $. On the other hand,
many trainable parameters lead to increasing complexity of the optimization
problem (\ref{RF_Att_49}) and to the possible overfitting of the whole RF.
\end{remark}

\section{Modifications of the two-level attention-based random forest}

Different configurations of LARF produce a set of models which depend on
trainable parameters of the two-level attention and its implementation. A
classification of models and their notations are shown in Table
\ref{t:RF_Att_L_models}. In order to explain the classification, two subsets
of the attention parameters should be considered:

\begin{enumerate}
\item Parameters produced by contamination probability distributions of the
Huber's $\epsilon$-contamination model in the form of a vector $\mathbf{w}$
whose length coincides with the number of trees.

\item Parameters $\epsilon_{1},...,\epsilon_{M}$ of contamination in the
mixture of $M$ the Huber's contamination models, which define imprecision of
the mixture model.
\end{enumerate}

The following models can be constructed depending on trainable parameters and
on using the \textquotedblleft leaf\textquotedblright\ attention, i.e., the
two-level attention mechanism:

\begin{itemize}
\item \textbf{$\epsilon$-ARF}: the attention-based forest with learning
$\epsilon$ as a attention parameter, but without training vector $\mathbf{w}$,
i.e., $w_{k}=1/T$, $k=1,...,T$, and without the \textquotedblleft
leaf\textquotedblright\ attention;

\item $\mathbf{w}$\textbf{-ARF}: the attention-based forest with learning
vector $\mathbf{w}$ as the attention parameters and without the
\textquotedblleft leaf\textquotedblright\ attention;

\item \textbf{$\epsilon$-LARF}: the attention-based forest with learning
$\epsilon$ as a attention parameter, but without training vector $\mathbf{w}$
and with the \textquotedblleft leaf\textquotedblright\ attention, i.e., by
using the two-level attention mechanism;

\item $\mathbf{w}$\textbf{-LARF}: the attention-based forest with learning
vector $\mathbf{w}$ as the attention parameters and with the \textquotedblleft
leaf\textquotedblright\ attention, i.e., by using the two-level attention mechanism;

\item $\epsilon$-$\mathbf{w}$\textbf{-ARF:} the attention-based forest with
learning vector $\mathbf{w}$ and the parameter $\epsilon$ as the attention
parameters and without the \textquotedblleft leaf\textquotedblright\ attention;

\item $\epsilon$-$\mathbf{w}$\textbf{-LARF:} the attention-based forest with
learning vector $\mathbf{w}$ and the parameter $\epsilon$ as the attention
parameters and with the \textquotedblleft leaf\textquotedblright\ attention;

\item $\epsilon M$-\textbf{ARF:} the attention-based forest with learning
parameters $\epsilon_{1},...,\epsilon_{M}$ as the attention parameters, with
$w_{k}=1/T$, $k=1,...,T$, and without the \textquotedblleft
leaf\textquotedblright\ attention;

\item $\epsilon M$-\textbf{LARF:} the attention-based forest with learning
parameters $\epsilon_{1},...,\epsilon_{M}$ as the attention parameters, with
$w_{k}=1/T$, $k=1,...,T$, and with the \textquotedblleft
leaf\textquotedblright\ attention;

\item $\epsilon M$-$\mathbf{w}$-\textbf{ARF:} the attention-based forest with
learning vector $\mathbf{w}$ and the parameters $\epsilon_{1},...,\epsilon
_{M}$ as the attention parameters and without the \textquotedblleft
leaf\textquotedblright\ attention;

\item $\epsilon M$-$\mathbf{w}$\textbf{-LARF:} the attention-based forest with
learning vector $\mathbf{w}$ and the parameters $\epsilon_{1},...,\epsilon
_{M}$ as the attention parameters and with the \textquotedblleft
leaf\textquotedblright\ attention.
\end{itemize}

Models $\epsilon$-ARF, $\epsilon$-LARF, $\epsilon$-$\mathbf{w}$-ARF,
$\epsilon$-$\mathbf{w}$-LARF are not presented in Table
\ref{t:RF_Att_L_models} because they are special cases of models $\epsilon
M$-ARF, $\epsilon M$-LARF, $\epsilon M$-$\mathbf{w}$-ARF, $\epsilon
M$-$\mathbf{w}$-LARF, respectively, by $M=1$.%

\begin{table}[tbp] \centering
\caption{Classification of the attention-based RF models proposed and studied in the paper}%
\begin{tabular}
[c]{ccccc}\hline
& \multicolumn{2}{c}{Tuning $\epsilon$} & \multicolumn{2}{c}{Trainable
$\epsilon_{1},...,\epsilon_{M}$}\\\hline
& fixed $\mathbf{w}$ & trainable $\mathbf{w}$ & fixed $\mathbf{w}$ & trainable
$\mathbf{w}$\\\hline
Without the \textquotedblleft leaf\textquotedblright\ attention & - &
$\mathbf{w}$-ARF & $\epsilon M$-ARF & $\epsilon M$-$\mathbf{w}$-ARF\\\hline
With the \textquotedblleft leaf\textquotedblright\ attention & - &
$\mathbf{w}$-LARF & $\epsilon M$-LARF & $\epsilon M$-$\mathbf{w}$-LARF\\\hline
\end{tabular}
\label{t:RF_Att_L_models}%
\end{table}%

\subsection{Huber's contamination model and the basic two-level attention}

In order to simplify optimization problem (\ref{RF_Att_50}) and to effectively
solve it, we propose to represent the attention weights $\alpha\left(
\mathbf{x},\mathbf{A}_{k}(\mathbf{x)},\mathbf{w}\right)  $ by using the
Huber's $\epsilon$-contamination model \cite{Huber81}. The idea to represent
the attention weight by means of the $\epsilon$-contamination model has been
proposed in \cite{Utkin-Konstantinov-22}. We use this idea to incorporate the
$\epsilon$-contamination model into optimization problem (\ref{RF_Att_50}) and
to construct first modifications of LARF.

Let us give a brief introduction into the Huber's $\epsilon$-contamination
model. The model considers a set of probability distributions of the form
$F(\mathbf{x})=(1-\epsilon)\cdot P(\mathbf{x})+\epsilon\cdot R$. Here
$P(\mathbf{x})=(p_{1}(\mathbf{x}),...,p_{T}(\mathbf{x}))$ is a discrete
probability distribution contaminated by another probability distribution
denoted $R=(r_{1},...,r_{T})$ which can be arbitrary in the unit simplex
having dimension $T$. It is important to note that the distribution $P$
depends on the feature vector $\mathbf{x}$, i.e., it is different for every
vector $\mathbf{x}$, whereas the distribution $R$ does not depend on
$\mathbf{x}$. Contamination parameter $\epsilon\in\lbrack0,1]$ controls the
impact of the contamination probability distribution $R$ on the distribution
$P(\mathbf{x})$. Since the distribution $R$ can be arbitrary, then the set the
distributions $F$ forms a subset of the unit simplex such that its size
depends on parameter $\epsilon$. If $\epsilon=0$, then the subset of
distributions $F$ is reduced to the single distribution $P(\mathbf{x})$. In
case of $\epsilon=1$, the set of $F(\mathbf{x})$ is the whole unit simplex.

Following the common definition of the attention weights through the softmax
operation with the parameter $\tau$, we propose to define each probability in
$P(\mathbf{x})$ as
\[
p_{k}(\mathbf{x})=\sigma\left(  -\left\Vert \mathbf{x}-\mathbf{A}%
_{k}(\mathbf{x)}\right\Vert ^{2}/\tau\right)  .
\]

This implies that distribution $P(\mathbf{x})$ characterizes how the feature
vector $\mathbf{x}$ is far from the vector $\mathbf{A}_{k}(\mathbf{x)}$ in all
trees of the RF. Let us suppose that the probability distribution $R$ is the
vector of trainable parameters $\mathbf{w}$. The idea is to train parameters
$\mathbf{w}$ to achieve the best accuracy of the RF. After substituting the
softmax operation into the attention weight $\alpha$, we get:
\begin{equation}
\alpha\left(  \mathbf{x},\mathbf{A}_{k}(\mathbf{x)},\mathbf{w}\right)
=(1-\epsilon)\cdot\text{$\sigma$}\left(  -\left\Vert \mathbf{x}-\mathbf{A}%
_{k}(\mathbf{x)}\right\Vert ^{2}/\tau\right)  +\epsilon\cdot w_{k}%
,\ k=1,...,T. \label{RSF_Att_90}%
\end{equation}

One can see from (\ref{RSF_Att_90}) that the attention weight is linearly
depends on trainable parameters $\mathbf{w}=(w_{1},...,w_{T})$. It is
important to note that the attention weight assigned to the $k$-th tree
depends only on the $k$-th parameter $w_{k}$, but not on other elements of
vector $\mathbf{w}$. Parameter $\epsilon$ is a tuning parameter determined by
means of the standard validation procedure. It should be noted that elements
of vector $\mathbf{w}$ are probabilities. Hence, there holds
\begin{equation}
\sum_{k=1}^{T}w_{k}=1.\ w_{k}\geq0,\ k=1,...,T. \label{RF_Att_L_55}%
\end{equation}

This implies that set $\mathcal{W}$ is the unit simplex of the dimension $T$.

Let us return to the attention weight $\mu\left(  \mathbf{x},\mathbf{x}%
_{j}\right)  $ of the first level. The attention is non-parametric at the
first level, therefore, the attention weight can be defined in the standard
way by using the Gaussian kernel or the softmax operation with parameter
$\tau_{0}$, i.e., there holds
\begin{equation}
\mu\left(  \mathbf{x},\mathbf{x}_{j}\right)  =\sigma\left(  -\left\Vert
\mathbf{x}-\mathbf{x}_{j}\right\Vert ^{2}/\tau_{0}\right)  .
\end{equation}

Finally, we can rewrite the loss function (\ref{RF_Att_50}) by taking into
account the above definitions of the attention weights as follows:%

\begin{align}
&  \min_{\mathbf{w\in}\mathcal{W}}\sum_{s=1}^{n}L\left(  \tilde{y}%
(\mathbf{x}_{s}),y_{s},\mathbf{w}\right) \nonumber\\
&  =\min_{\mathbf{w\in}\mathcal{W}}\sum_{s=1}^{n}\left(  y_{s}-\sum_{k=1}%
^{T}\left(  (1-\epsilon)C_{k}(\mathbf{x}_{s})-\epsilon D_{k}(\mathbf{x}%
_{s})w_{k}\right)  \right)  ^{2}, \label{RF_Att_L_60}%
\end{align}
where
\begin{align}
C_{k}(\mathbf{x}_{s})  &  =\sum\nolimits_{l\in\mathcal{J}_{k}(\mathbf{x}_{s}%
)}y_{l}\cdot\sigma\left(  -\frac{\left\Vert \mathbf{x}_{s}-\mathbf{x}%
_{l}\right\Vert ^{2}}{\tau_{0}}\right) \nonumber\\
&  \times\text{$\sigma$}\left(  -\frac{\left\Vert \mathbf{x}-\sum
\nolimits_{i\in\mathcal{J}_{k}(\mathbf{x}_{s})}\mathbf{x}_{i}\cdot
\sigma\left(  -\left\Vert \mathbf{x}_{s}-\mathbf{x}_{i}\right\Vert ^{2}%
/\tau_{0}\right)  \right\Vert ^{2}}{\tau}\right)  , \label{RF_Att_L_62}%
\end{align}%
\begin{equation}
D_{k}(\mathbf{x}_{s})=\sum\nolimits_{j\in\mathcal{J}_{k}(\mathbf{x}_{s})}%
y_{j}\cdot\sigma\left(  -\frac{\left\Vert \mathbf{x}_{s}-\mathbf{x}%
_{j}\right\Vert ^{2}}{\tau_{0}}\right)  . \label{RF_Att_L_64}%
\end{equation}

One can see from the above that $C_{k}(\mathbf{x}_{s})$ and $D_{k}%
(\mathbf{x}_{s})$ do not depend on parameters $\mathbf{w}$. Therefore, the
objective function (\ref{RF_Att_L_60}) jointly with the simple constraints
$\mathbf{w}\in\mathcal{W}$ or (\ref{RF_Att_L_55}) is the standard quadratic
optimization problem which can be solved by means of many available efficient
algorithms. The corresponding model is called $\mathbf{w}$\textbf{-}LARF. The
notation means that trainable parameters are $\mathbf{w}$. The same model
without \textquotedblleft leaf\textquotedblright\ attention is denoted as
$\mathbf{w}$\textbf{-}ARF. It coincides with the model $\epsilon$-ABRF
proposed in \cite{Utkin-Konstantinov-22}.

It should be noted that the problem (\ref{RF_Att_L_60}) is similar to the
optimization problem stated in
\cite{Konstantinov-Utkin-22d,Utkin-Konstantinov-22}. However, it turns out
that the addition of the \textquotedblleft leaf\textquotedblright\ attention
significantly improves the RF as it will be shown by many numerical
experiments with real data.

\subsection{Models with trainable contamination parameter $\epsilon$}

One of the important contributions to the work, which makes the proposed model
different from the model presented in
\cite{Konstantinov-Utkin-22d,Utkin-Konstantinov-22}, is the idea to learn the
contamination parameter $\epsilon$ jointly with parameters $\mathbf{w}$.
However, this idea leads to a complex optimization problem such that
gradient-based algorithms have to be used. In order to avoid using these
algorithms and to get a simple optimization problem, we consider two ways. The
first way is just to assign the same value $1/T$ to all parameters $w_{k}$.
Then the optimization problem (\ref{RF_Att_L_60}) can be rewritten as%
\begin{equation}
\min_{0\leq\epsilon\leq1}\sum_{s=1}^{n}\left(  y_{s}-\sum_{k=1}^{T}\left(
(1-\epsilon)C_{k}(\mathbf{x}_{s})-\epsilon D_{k}(\mathbf{x}_{s})\frac{1}%
{T}\right)  \right)  ^{2}.
\end{equation}

We get a simple quadratic optimization problem with one variable $\epsilon$.
Let us call the corresponding model as \textbf{$\epsilon$}-LARF\textbf{. }The
notation means that the trainable parameter is $\epsilon$. The same model
without \textquotedblleft leaf\textquotedblright\ attention is denoted as
$\epsilon$-ARF.

Another way is to introduce new variables $\gamma_{k}=\epsilon w_{k}$,
$k=1,...,T$. Then the optimization problem (\ref{RF_Att_L_60}) can be
rewritten as
\begin{equation}
\min_{\gamma_{1},...,\gamma_{T},\epsilon}\sum_{s=1}^{n}\left(  y_{s}%
-(1-\epsilon)\sum_{k=1}^{T}C_{k}(\mathbf{x}_{s})-\sum_{k=1}^{T}\gamma_{k}%
D_{k}(\mathbf{x}_{s})\right)  ^{2}, \label{RF_Att_L_70}%
\end{equation}
subject to%
\begin{equation}
\sum_{k=1}^{T}\gamma_{k}=\epsilon,\ \gamma_{k}\geq0,\ k=1,...,T,
\label{RF_Att_L_71}%
\end{equation}
\begin{equation}
0+\varsigma\leq\epsilon\leq1. \label{RF_Att_L_72}%
\end{equation}

We again get the quadratic optimization problem with new optimization
variables $\gamma_{1},...,\gamma_{T},\epsilon$ and linear constraints. The
parameter $\varsigma$ takes a small value to avoid the case $\epsilon=0$. The
corresponding model is denoted $\epsilon$-$\mathbf{w}$\textbf{-}LARF. The
notation means that trainable parameters are $\mathbf{w}$ and $\epsilon$. The
same model without \textquotedblleft leaf\textquotedblright\ attention is
denoted as $\epsilon$-$\mathbf{w}$\textbf{-}ARF.

\subsection{Mixture of contamination models}

Another important contribution is an attempt to search for an optimal value of
the temperature parameter $\tau$ in (\ref{RSF_Att_90}) or in
(\ref{RF_Att_L_62}). We propose an approximate approach which can
significantly improve the model. Let us introduce a finite set $\{\tau
_{1},...,\tau_{M}\}$ of $M$ values of parameter $\tau$. Here $M$ can be
regarded as a tuning integer parameter which impact on the number of all
training parameters. Large values of $M$ may lead to a large number of
training parameters and the corresponding overfitting. Small values of $M$ may
lead to an inexact approximation of $\tau$.

Before considering how the optimization problem can be rewritten taking into
account the above, we again return to the attention weight $\alpha$ in
(\ref{RSF_Att_90}) and represent it as follows:%
\begin{equation}
\alpha\left(  \mathbf{x},\mathbf{A}_{k}(\mathbf{x)},w_{k}\right)  =\frac{1}%
{M}\sum_{j=1}^{M}\alpha_{j}\left(  \mathbf{x},\mathbf{A}_{k}(\mathbf{x)}%
,w_{k}\right)  ,
\end{equation}
where
\begin{equation}
\alpha_{j}\left(  \mathbf{x},\mathbf{A}_{k}(\mathbf{x)},w_{k}\right)
=(1-\epsilon_{j})\text{$\sigma$}\left(  -\frac{\left\Vert \mathbf{x}%
-\mathbf{A}_{k}(\mathbf{x)}\right\Vert ^{2}}{\tau_{j}}\right)  +\epsilon
_{j}w_{k}.
\end{equation}

\begin{figure}
[ptb]
\begin{center}
\includegraphics[
height=1.8258in,
width=4.166in
]%
{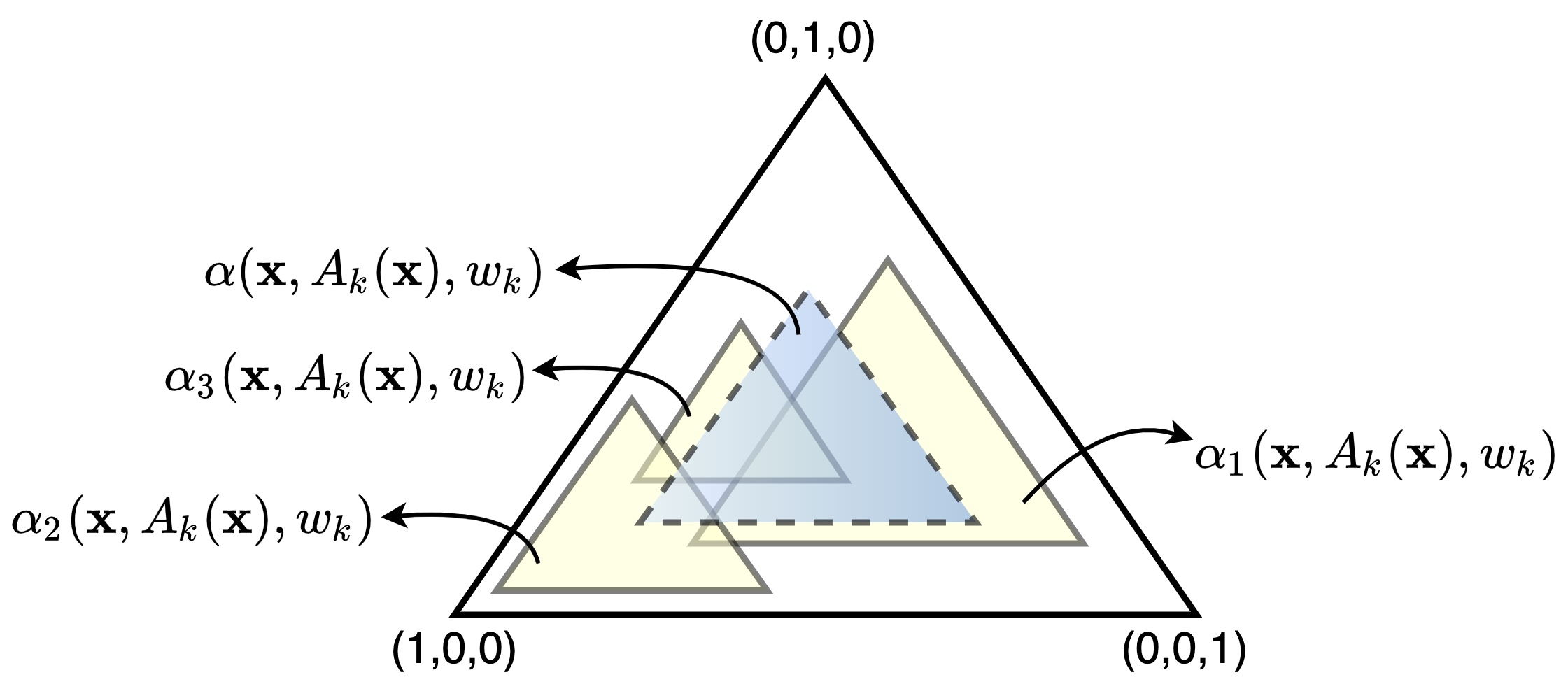}%
\caption{The unit simplex of possible attention weights which includes small
simplices corresponding to three contamination models with different centers
and different contamination parameters $\epsilon$ and the \textquotedblleft
mean\textquotedblright\ simplex (with dashed sides) which defines the set of
final attention weights}%
\label{f:simplex_att_l}%
\end{center}
\end{figure}

We have a mixture of $M$ contamination models with different contamination
parameters $\epsilon_{j}$. It is obvious that the sum of new weights $\alpha$
over $k=1,...,T$ is $1$ because the sum of each $\alpha_{j}\left(
\mathbf{x},\mathbf{A}_{k}(\mathbf{x)},w_{k}\right)  $ over $k=1,...,T$ is also
$1$. Each $\alpha_{j}\left(  \mathbf{x},\mathbf{A}_{k}(\mathbf{x)}%
,w_{k}\right)  $ forms a small simplex such that its center is defined by
$\tau_{j}$ and its size is defined by $\tau_{j}$. The corresponding sets of
possible attention weights are depicted in Fig. \ref{f:simplex_att_l} where
the unit simplex by $T=3$ includes small simplices corresponding to three
($M=3$) contamination models $\alpha_{j}\left(  \mathbf{x},\mathbf{A}%
_{k}(\mathbf{x)},w_{k}\right)  $, $j=1,2,3$, with different centers and
different contamination parameters $\epsilon_{1},\epsilon_{2},\epsilon_{3}$.
The \textquotedblleft mean\textquotedblright\ simplex of weights
$\alpha\left(  \mathbf{x},\mathbf{A}_{k}(\mathbf{x)},w_{k}\right)  $ is
depicted by using dashed sides. Parameters $\mathbf{w}$ are optimized such
that the attention weights will be located in the \textquotedblleft
mean\textquotedblright\ simplex.%

Let us prove that the resulting \textquotedblleft mean\textquotedblright%
\ model represents the $\epsilon$-contamination model with a contamination
parameter $\epsilon$. Denote
\begin{equation}
P_{j}=\sigma\left(  -\frac{\left\Vert \mathbf{x}-\mathbf{A}_{k}(\mathbf{x)}%
\right\Vert ^{2}}{\tau_{j}}\right)  .
\end{equation}

Here $P_{j}$ is a probability distribution, i.e., $P_{j}=(p_{j}^{(1)}%
,...,p_{j}^{(T)})$. Then we can write
\begin{align}
\alpha\left(  \mathbf{x},\mathbf{A}_{k}(\mathbf{x)},w_{k}\right)   &
=\frac{1}{M}\sum_{j=1}^{M}(1-\epsilon_{j})P_{j}+\frac{1}{M}\sum_{j=1}%
^{M}\epsilon_{j}w_{k}\nonumber\\
&  =\frac{1}{M}\sum_{j=1}^{M}(1-\epsilon_{j})P_{j}+\epsilon w_{k},
\end{align}
where
\begin{equation}
\epsilon=\frac{1}{M}\sum_{j=1}^{M}\epsilon_{j}. \label{RF_Att_L_67}%
\end{equation}

Suppose that there exists a probability distribution $Q=(q^{(1)},...,q^{(T)})$
such that there holds
\begin{equation}
\frac{1}{M}\sum_{j=1}^{M}(1-\epsilon_{j})P_{j}=(1-\epsilon)Q.
\label{RF_Att_L_68}%
\end{equation}

If we prove that the probability distribution $Q$ exists, then the resulting
\textquotedblleft mean\textquotedblright\ model is the $\epsilon
$-contamination model. Let us find sums of the left and the right sides of
(\ref{RF_Att_L_68}) over $i=1,...,T$. Hence, we get
\begin{equation}
\frac{1}{M}\sum_{j=1}^{M}(1-\epsilon_{j})\sum_{i=1}^{T}p_{j}^{(i)}%
=(1-\epsilon)\sum_{i=1}^{T}q^{(i)}. \label{RF_Att_L_69}%
\end{equation}

Substituting (\ref{RF_Att_L_67}) into (\ref{RF_Att_L_69}), we get
\[
\sum_{i=1}^{T}q^{(i)}=1,
\]
as was to be proved.

The introduced mixture of the contamination models can be regarded as a
multi-head attention to some extent where every \textquotedblleft
head\textquotedblright\ is produced by using a certain parameter $\tau_{j}$.

Let us represent the softmax operation in (\ref{RSF_Att_90}) jointly with the
factor $(1-\epsilon)$ as follows:
\begin{equation}
(1-\epsilon)\cdot\text{$\sigma$}\left(  \frac{-\left\Vert \mathbf{x}%
-\mathbf{A}_{k}(\mathbf{x)}\right\Vert ^{2}}{\tau}\right)  =\frac{1}{M}%
\sum_{j=1}^{M}(1-\epsilon_{j})\cdot\text{$\sigma$}\left(  \frac{-\left\Vert
\mathbf{x}-\mathbf{A}_{k}(\mathbf{x)}\right\Vert ^{2}}{\tau_{j}}\right)  .
\label{RF_Att_L_73}%
\end{equation}

It can be seen from (\ref{RF_Att_L_73}) that new parameters $\epsilon
_{1},...,\epsilon_{M}$ along with $\tau_{1},...,\tau_{M}$ are introduced in
place of $\epsilon$ and $\tau$, respectively. Term $(1-\epsilon)\cdot$
$C_{k}(\mathbf{x}_{s})$ in (\ref{RF_Att_L_60}) and (\ref{RF_Att_L_62}) is
replaced with the following terms:
\begin{equation}
\frac{1}{M}\sum_{j=1}^{M}(1-\epsilon_{j})C_{k}^{(j)}(\mathbf{x}_{s}),
\end{equation}
where%
\begin{align}
C_{k}^{(j)}(\mathbf{x}_{s})  &  =\sum\nolimits_{l\in\mathcal{J}_{k}%
(\mathbf{x}_{s})}y_{l}\cdot\sigma\left(  -\frac{\left\Vert \mathbf{x}%
_{s}-\mathbf{x}_{l}\right\Vert ^{2}}{\tau_{0}}\right) \nonumber\\
&  \times\text{$\sigma$}\left(  -\frac{\left\Vert \mathbf{x}_{s}%
-\sum\nolimits_{i\in\mathcal{J}_{k}(\mathbf{x}_{s})}\mathbf{x}_{i}\mu\left(
\mathbf{x}_{s},\mathbf{x}_{i}\right)  \right\Vert ^{2}}{\tau_{j}}\right)  .
\end{align}

Finally, we get the following optimization problem:%
\begin{align}
&  \min_{\gamma_{1},...,\gamma_{T},\epsilon}\sum_{s=1}^{n}\left(  y_{s}%
-\frac{1}{M}\sum_{j=1}^{M}(1-\epsilon_{j})\sum_{k=1}^{T}C_{k}^{(j)}%
(\mathbf{x}_{s})\right. \nonumber\\
&  \left.  -\frac{1}{M}\sum_{j=1}^{M}\epsilon_{j}\sum_{k=1}^{T}w_{k}%
D_{k}(\mathbf{x}_{s})\right)  ^{2},
\end{align}
subject to%
\[
\sum_{k=1}^{T}w_{k}=1,~w_{k}\geq0,~k=1,...,T.
\]

Introduce new variables $\gamma_{k}^{(j)}=w_{k}\epsilon_{j}$, $k=1,...,T$,
$j=1,...,M$. Hence we can write the optimization problem with new $M\cdot T+M$
variables as%
\begin{align}
&  \min_{\gamma_{1},...,\gamma_{T},\epsilon}\sum_{s=1}^{n}\left(  y_{s}%
-\frac{1}{M}\sum_{j=1}^{M}(1-\epsilon_{j})\sum_{k=1}^{T}C_{k}^{(j)}%
(\mathbf{x}_{s})\right. \nonumber\\
&  \left.  -\frac{1}{M}\sum_{j=1}^{M}\sum_{k=1}^{T}\gamma_{k}^{(j)}%
D_{k}(\mathbf{x}_{s})\right)  ^{2},
\end{align}
subject to
\begin{equation}
\sum_{k=1}^{T}\gamma_{k}^{(j)}=\epsilon_{j},\ \gamma_{k}^{(j)}\geq
0,\ k=1,...,T,~,j=1,...,M,
\end{equation}%
\begin{equation}
0+\varsigma\leq\epsilon_{j}\leq1.
\end{equation}

We again get the quadratic optimization problem with linear constraints. The
corresponding model will be denoted as $\epsilon M$-$\mathbf{w}$-ARF or
$\epsilon M$-$\mathbf{w}$-LARF depending on\textbf{ }the applying the
\textquotedblleft leaf\textquotedblright\ attention. The notation means that
trainable parameters are $\mathbf{w}$ and $\epsilon_{1},...,\epsilon_{M}$. We
also will use the same model, but with condition $w_{k}=1/T$ for all
$k=1,...,T$. The corresponding models are denoted as $\epsilon M$-ARF or
$\epsilon M$-LARF.

\section{Numerical experiments}

Let is introduce notation for different models of the attention-based RFs.

\begin{enumerate}
\item \textbf{RF }(\textbf{ERT}): the original RF (the ERT) without applying
the attention mechanisms;

\item \textbf{ARF (LARF)}: the attention-based forest having the following
modifications: $\epsilon M$-\textbf{ARF}, $\epsilon M$-\textbf{LARF},
$\epsilon M$-$\mathbf{w}$-\textbf{ARF}, $\epsilon M$-$\mathbf{w}%
$\textbf{-LARF}.
\end{enumerate}

In all experiments, RFs as well as ERTs consist of $100$ trees. For selecting
the best tuning parameters in numerical experiments, a 3-fold cross-validation
on the training set consisting of $n_{\text{tr}}=4n/5$ examples with $100$
repetitions is performed. The search for the best parameter $\tau_{0}$ is
carried out by considering all its values in a predefined grid. A
cross-validation procedure is subsequently used to select their appropriate
values. The testing set for computing the accuracy measures consists of
$n_{\text{test}}=n/5$ examples. In order to get desirable estimates of vectors
$\mathbf{A}_{k}(\mathbf{x})$ and $B_{k}(\mathbf{x})$, all trees in experiments
are trained such that at least $10$ examples fall into every leaf of trees.

We do not consider models \textbf{$\epsilon$}-ARF, $\mathbf{w}$-ARF,
\textbf{$\epsilon$}-LARF, $\mathbf{w}$-LARF because they can be regarded as
special cases of the corresponding models \textbf{$\epsilon$}$M$-ARF,
$\epsilon M$-$\mathbf{w}$-ARF, \textbf{$\epsilon$}$M$-LARF, $\epsilon
M$-$\mathbf{w}$-LARF when $M=1$. The value of $M$ is taken $10$. Set
$\{\tau_{1},...,\tau_{M}\}$ of the softmax operation parameters is defined as
$\left\{  10^{-\left\lfloor M/2\right\rfloor },10^{-\left\lfloor
M/2\right\rfloor +1},...,10^{0},...,10^{\left\lfloor M/2\right\rfloor
-1},10^{\left\lfloor M/2\right\rfloor }\right\}  $. In particular, if $M=1$,
then the set of $\tau$ consists of one element $\tau=1$. Parameter of the
first-level attention in \textquotedblleft leaf\textquotedblright\ $\tau_{0}$
is taken equal to $1$.

Numerical results are presented in tables where the best results are shown in
bold. The coefficient of determination denoted $R^{2}$ and the mean absolute
error (MAE) are used for the regression evaluation. The greater the value of
the coefficient of determination and the smaller the MAE, the better results
we get.

The proposed approach is studied by applying datasets which are taken from
open sources. The dataset Diabetes is downloaded from the R Packages; datasets
Friedman 1, 2 and 3 are taken at site:
https://www.stat.berkeley.edu/\symbol{126}breiman/bagging.pdf; datasets
Regression and Sparse are taken from package \textquotedblleft
Scikit-Learn\textquotedblright; datasets Wine Red, Boston Housing, Concrete,
Yacht Hydrodynamics, Airfoil can be found in the UCI Machine Learning
Repository \cite{Dua:2019}. These datasets with their numbers of features $m$
and numbers of examples $n$ are given in Table \ref{t:regres_datasets}. A more
detailed information can be found from the aforementioned data resources.%

\begin{table}[tbp] \centering
\caption{A brief introduction about the regression data sets}%
\begin{tabular}
[c]{cccc}\hline
Data set & Abbreviation & $m$ & $n$\\\hline
Diabetes & Diabetes & $10$ & $442$\\\hline
Friedman 1 & Friedman 1 & $10$ & $100$\\\hline
Friedman 2 & Friedman 2 & $4$ & $100$\\\hline
Friedman 3 & Friedman 3 & $4$ & $100$\\\hline
Scikit-Learn Regression & Regression & $100$ & $100$\\\hline
Scikit-Learn Sparse Uncorrelated & Sparse & $10$ & $100$\\\hline
UCI Wine red & Wine & $11$ & $1599$\\\hline
UCI Boston Housing & Boston & $13$ & $506$\\\hline
UCI Concrete & Concrete & $8$ & $1030$\\\hline
UCI Yacht Hydrodynamics & Yacht & $6$ & $308$\\\hline
UCI Airfoil & Airfoil & $5$ & $1503$\\\hline
\end{tabular}
\label{t:regres_datasets}%
\end{table}%

Values of the measure $R^{2}$ for several models, including RF, $\epsilon
M$-$\mathbf{w}$-ARF, $\epsilon M$-$\mathbf{w}$-LARF, $\epsilon M$-ARF and
$\epsilon M$-LARF, are shown in Table \ref{t:RF_Att_L_1}. The results are
obtained by training the RF. Optimal values of $\tau_{0}$ are also given in
the table. It can be seen from Table \ref{t:RF_Att_L_1} that $\epsilon
M$-$\mathbf{w}$-LARF outperforms all models for most datasets. Moreover, one
can see from Table \ref{t:RF_Att_L_1} that the two-level attention models
($\epsilon M$-$\mathbf{w}$-LARF and $\epsilon M$-LARF) provide better results
than models which do not use the \textquotedblleft leaf\textquotedblright%
\ attention ($\epsilon M$-$\mathbf{w}$-ARF and $\epsilon M$-ARF). It should be
also noted that all attention-based models outperform the original RF. The
same relationship between the models takes place for another accuracy measure
(MAE). It is clearly shown in Table \ref{t:RF_Att_L_2}.%

\begin{table}[tbp] \centering
\caption{Values of $R^2$ for comparison of models based on the RF}%
\begin{tabular}
[c]{ccccccc}\hline
Data set & $\tau_{0}$ & RF & $\epsilon M$-$\mathbf{w}$-ARF & $\epsilon
M$-$\mathbf{w}$-LARF & $\epsilon M$-ARF & $\epsilon M$-LARF\\\hline
Diabetes & $0.01$ & $0.416$ & $0.419$ & $\mathbf{0}$\textbf{$.$}$\mathbf{434}$
& $0.425$ & $0.426$\\\hline
Friedman 1 & $1$ & $0.459$ & $0.470$ & $\mathbf{0}$\textbf{$.$}$\mathbf{524}$
& $0.438$ & $0.472$\\\hline
Friedman 2 & $1$ & $0.841$ & $0.887$ & $\mathbf{0}$\textbf{$.$}$\mathbf{933}$
& $0.886$ & $0.916$\\\hline
Friedman 3 & $1$ & $0.625$ & $0.708$ & $\mathbf{0}$\textbf{$.$}$\mathbf{749}$
& $0.675$ & $0.704$\\\hline
Airfoil & $100$ & $0.823$ & $0.844$ & $0.914$ & $0.822$ & $\mathbf{0}%
$\textbf{$.$}$\mathbf{917}$\\\hline
Boston & $1$ & $0.814$ & $0.820$ & $\mathbf{0}$\textbf{$.$}$\mathbf{870}$ &
$0.819$ & $0.856$\\\hline
Concrete & $10$ & $0.845$ & $0.857$ & $\mathbf{0}$\textbf{$.$}$\mathbf{896}$ &
$0.844$ & $\mathbf{0}$\textbf{$.$}$\mathbf{896}$\\\hline
Wine & $1$ & $0.433$ & $0.421$ & $\mathbf{0}$\textbf{$.$}$\mathbf{481}$ &
$0.421$ & $0.477$\\\hline
Yacht & $0.1$ & $0.981$ & $0.989$ & $\mathbf{0}$\textbf{$.$}$\mathbf{993}$ &
$0.981$ & $0.982$\\\hline
Regression & $0.1$ & $0.380$ & $0.434$ & $\mathbf{0}$\textbf{$.$}%
$\mathbf{455}$ & $0.361$ & $0.409$\\\hline
Sparse & $1$ & $0.470$ & $0.489$ & $\mathbf{0}$\textbf{$.$}$\mathbf{641}$ &
$0.535$ & $0.630$\\\hline
\end{tabular}
\label{t:RF_Att_L_1}%
\end{table}%
%

\begin{table}[tbp] \centering
\caption{Values of MAE for comparison of models based on the RF}%
\begin{tabular}
[c]{cccccc}\hline
Data set & RF & $\epsilon M$-$\mathbf{w}$-ARF & $\epsilon M$-$\mathbf{w}%
$-LARF & $\epsilon M$-ARF & $\epsilon M$-LARF\\\hline
Diabetes & $44.92$ & $44.95$ & $\mathbf{44}$\textbf{$.$}$\mathbf{61}$ &
$44.79$ & $44.81$\\\hline
Friedman 1 & $2.540$ & $2.545$ & $\mathbf{2}$\textbf{$.$}$\mathbf{411}$ &
$2.595$ & $2.473$\\\hline
Friedman 2 & $111.7$ & $95.29$ & $\mathbf{72}$\textbf{$.$}$\mathbf{71}$ &
$92.44$ & $74.24$\\\hline
Friedman 3 & $0.154$ & $0.130$ & $0.135$ & $0.144$ & $\mathbf{0}$\textbf{$.$%
}$\mathbf{129}$\\\hline
Airfoil & $2.203$ & $2.065$ & $1.451$ & $2.217$ & $\mathbf{1}$\textbf{$.$%
}$\mathbf{416}$\\\hline
Boston & $2.539$ & $2.538$ & $\mathbf{2}$\textbf{$.$}$\mathbf{148}$ & $2.489$
& $2.217$\\\hline
Concrete & $4.834$ & $4.676$ & $\mathbf{3}$\textbf{$.$}$\mathbf{496}$ &
$4.883$ & $3.615$\\\hline
Wine & $0.451$ & $0.459$ & $\mathbf{0}$\textbf{$.$}$\mathbf{411}$ & $0.461$ &
$0.417$\\\hline
Yacht & $1.004$ & $0.787$ & $\mathbf{0}$\textbf{$.$}$\mathbf{611}$ & $1.004$ &
$0.971$\\\hline
Regression & $109.1$ & $103.6$ & $\mathbf{101}$\textbf{$.$}$\mathbf{3}$ &
$111.2$ & $105.8$\\\hline
Sparse & $1.908$ & $1.871$ & $\mathbf{1}$\textbf{$.$}$\mathbf{528}$ & $1.772$
& $1.543$\\\hline
\end{tabular}
\label{t:RF_Att_L_2}%
\end{table}%

Another important question is how the attention-based models perform when the
ERT is used. The corresponding values of $R^{2}$ and MAE are shown in Tables
\ref{t:RF_Att_L_3} and \ref{t:RF_Att_L_4}, respectively. Table
\ref{t:RF_Att_L_3} also contains optimal values $\tau_{0}$. In contrast to the
case of using the RF, it can be seen from the tables that $\epsilon M$-LARF
outperforms $\epsilon M$-$\mathbf{w}$-LARF for several models. It can be
explained by reducing the accuracy due to a larger number of training
parameters (parameters $\mathbf{w}$) and overfitting for small datasets. It is
also interesting to note that models based on ERTs provide better results than
models based on RFs. However, this improvement is not significant. This is
clearly seen from Table \ref{t:RF_Att_L_6} where the best results are
collected for models based on ERTs and RFs. One can see from Table
\ref{t:RF_Att_L_6} that results are identical for several datasets, namely,
for datasets Friedman 1, 2, 3, Concrete, Yacht. If to apply the $t$-test to
compare values of $R^{2}$ obtained for two models, then, according to
\cite{Demsar-2006}, the $t$-statistics is distributed in accordance with the
Student distribution with $11-1$ degrees of freedom ($11$ datasets). The
obtained p-value is $p=0.071$. We can conclude that the outperformance of the
ERT is not statistically significant because $p>0.05$.%

\begin{table}[tbp] \centering
\caption{Values of $R^2$ for comparison of models based on the ERT}%
\begin{tabular}
[c]{ccccccc}\hline
Data set & $\tau_{0}$ & ERT & $\epsilon M$-$\mathbf{w}$-ARF & $\epsilon
M$-$\mathbf{w}$-LARF & $\epsilon M$-ARF & $\epsilon M$-LARF\\\hline
Diabetes & $0.01$ & $0.438$ & $0.441$ & $0.434$ & $\mathbf{0}$\textbf{$.$%
}$\mathbf{471}$ & $0.444$\\\hline
Friedman 1 & $1$ & $0.471$ & $0.471$ & $\mathbf{0}$\textbf{$.$}$\mathbf{524}$
& $0.441$ & $0.495$\\\hline
Friedman 2 & $10$ & $0.813$ & $0.840$ & $\mathbf{0}$\textbf{$.$}$\mathbf{933}$
& $0.840$ & $0.919$\\\hline
Friedman 3 & $1$ & $0.570$ & $0.569$ & $\mathbf{0}$\textbf{$.$}$\mathbf{749}$
& $0.569$ & $0.637$\\\hline
Airfoil & $100$ & $0.802$ & $0.804$ & $\mathbf{0}$\textbf{$.$}$\mathbf{914}$ &
$0.804$ & $0.909$\\\hline
Boston & $10$ & $0.831$ & $0.834$ & $0.870$ & $0.834$ & $\mathbf{0}%
$\textbf{$.$}$\mathbf{882}$\\\hline
Concrete & $10$ & $0.839$ & $0.838$ & $\mathbf{0}$\textbf{$.$}$\mathbf{896}$ &
$0.838$ & $0.895$\\\hline
Wine & $1$ & $0.418$ & $0.418$ & $0.481$ & $0.418$ & $\mathbf{0}$\textbf{$.$%
}$\mathbf{486}$\\\hline
Yacht & $1$ & $0.988$ & $0.988$ & $\mathbf{0}$\textbf{$.$}$\mathbf{993}$ &
$0.988$ & $\mathbf{0}$\textbf{$.$}$\mathbf{993}$\\\hline
Regression & $0.1$ & $0.402$ & $0.429$ & $0.455$ & $0.429$ & $\mathbf{0}%
$\textbf{$.$}$\mathbf{464}$\\\hline
Sparse & $1$ & $0.452$ & $0.522$ & $0.641$ & $0.522$ & $\mathbf{0}$%
\textbf{$.$}$\mathbf{663}$\\\hline
\end{tabular}
\label{t:RF_Att_L_3}%
\end{table}%
%

\begin{table}[tbp] \centering
\caption{Values of MAE for comparison of models based on the ERT}%
\begin{tabular}
[c]{cccccc}\hline
Data set & ERT & $\mathbf{w}$-ARF & $\mathbf{w}$-LARF & $\epsilon M$-ARF &
$\epsilon M$-LARF\\\hline
Diabetes & $44.549$ & $44.271$ & $44.614$ & $44.271$ & $\mathbf{44}%
$\textbf{$.$}$\mathbf{21}$\\\hline
Friedman 1 & $2.502$ & $2.502$ & $2.411$ & $2.502$ & $\mathbf{2}$\textbf{$.$%
}$\mathbf{388}$\\\hline
Friedman 2 & $123.0$ & $113.7$ & $72.71$ & $113.7$ & $\mathbf{70}$\textbf{$.$%
}$\mathbf{48}$\\\hline
Friedman 3 & $0.179$ & $0.179$ & $\mathbf{0}$\textbf{$.$}$\mathbf{135}$ &
$0.179$ & $0.148$\\\hline
Airfoil & $2.370$ & $2.360$ & $\mathbf{1}$\textbf{$.$}$\mathbf{451}$ & $2.360$
& $1.471$\\\hline
Boston & $2.481$ & $2.451$ & $2.148$ & $2.451$ & $\mathbf{2}$\textbf{$.$%
}$\mathbf{023}$\\\hline
Concrete & $5.119$ & $5.124$ & $\mathbf{3}$\textbf{$.$}$\mathbf{496}$ &
$5.124$ & $3.659$\\\hline
Wine & $0.464$ & $0.464$ & $\mathbf{0}$\textbf{$.$}$\mathbf{411}$ & $0.464$ &
$0.412$\\\hline
Yacht & $0.824$ & $0.822$ & $\mathbf{0}$\textbf{$.$}$\mathbf{611}$ & $0.822$ &
$0.612$\\\hline
Regression & $106.3$ & $103.1$ & $101.3$ & $103.1$ & $\mathbf{100}$%
\textbf{$.$}$\mathbf{0}$\\\hline
Sparse & $1.994$ & $1.820$ & $1.528$ & $1.820$ & $\mathbf{1}$\textbf{$.$%
}$\mathbf{519}$\\\hline
\end{tabular}
\label{t:RF_Att_L_4}%
\end{table}%
%

\begin{table}[tbp] \centering
\caption{Comparison of the best results provided by models based on RFs and ERTs}%
\begin{tabular}
[c]{ccc}\hline
Data set & RF & ERT\\\hline
Diabetes & $0.434$ & $\mathbf{0}$\textbf{$.$}$\mathbf{471}$\\\hline
Friedman 1 & $\mathbf{0}$\textbf{$.$}$\mathbf{524}$ & $\mathbf{0}$\textbf{$.$%
}$\mathbf{524}$\\\hline
Friedman 2 & $\mathbf{0}$\textbf{$.$}$\mathbf{933}$ & $\mathbf{0}$\textbf{$.$%
}$\mathbf{933}$\\\hline
Friedman 3 & $\mathbf{0}$\textbf{$.$}$\mathbf{749}$ & $\mathbf{0}$\textbf{$.$%
}$\mathbf{749}$\\\hline
Airfoil & $\mathbf{0}$\textbf{$.$}$\mathbf{917}$ & $0.914$\\\hline
Boston & $0.870$ & $\mathbf{0}$\textbf{$.$}$\mathbf{882}$\\\hline
Concrete & $\mathbf{0}$\textbf{$.$}$\mathbf{896}$ & $\mathbf{0}$\textbf{$.$%
}$\mathbf{896}$\\\hline
Wine & $0.481$ & $\mathbf{0}$\textbf{$.$}$\mathbf{486}$\\\hline
Yacht & $\mathbf{0}$\textbf{$.$}$\mathbf{993}$ & $\mathbf{0}$\textbf{$.$%
}$\mathbf{993}$\\\hline
Regression & $0.455$ & $\mathbf{0}$\textbf{$.$}$\mathbf{464}$\\\hline
Sparse & $0.641$ & $\mathbf{0}$\textbf{$.$}$\mathbf{663}$\\\hline
\end{tabular}
\label{t:RF_Att_L_6}%
\end{table}%

It should be pointed out that the proposed models can be regarded as
extensions of the attention-based RF ($\epsilon$-ABRF) presented in
\cite{Utkin-Konstantinov-22}. Therefore, it is also interesting to compare the
two-level attention models with $\epsilon$-ABRF. Table \ref{t:RF_Att_L_7}
shows values of $R^{2}$ obtained by using $\epsilon$-ABRF and the best values
of proposed models when the RF and the ERT are used.

If to formally compare results presented in Table \ref{t:RF_Att_L_7} by
applying the $t$-tests in accordance with \cite{Demsar-2006}, then tests for
the proposed models and $\epsilon$-ABRF based on the RF and the ERT provide
p-values equal to $p=0.00067$ and $p=0.00029$, respectively. The tests
demonstrate the clear outperformance of the proposed models in comparison with
$\epsilon$-ABRF.%

\begin{table}[tbp] \centering
\caption{Comparison of $\epsilon $-ABRF and the proposed models by using the measure $R^2$ when RFs and ERTs are the basis}%
\begin{tabular}
[c]{ccccc}\hline
& \multicolumn{2}{c}{RF} & \multicolumn{2}{c}{ERT}\\\hline
Data set & $\epsilon$-ABRF & LARF & $\epsilon$-ABRF & LARF\\\hline
Diabetes & $0.424$ & $\mathbf{0.434}$ & $0.441$ & $\mathbf{0.471}$\\\hline
Friedman 1 & $0.470$ & $\mathbf{0.524}$ & $0.513$ & $\mathbf{0.524}$\\\hline
Friedman 2 & $0.877$ & $\mathbf{0.933}$ & $0.930$ & $\mathbf{0.933}$\\\hline
Friedman 3 & $0.686$ & $\mathbf{0.749}$ & $0.739$ & $\mathbf{0.749}$\\\hline
Airfoil & $0.843$ & $\mathbf{0.917}$ & $0.837$ & $\mathbf{0.914}$\\\hline
Boston & $0.823$ & $\mathbf{0.870}$ & $0.838$ & $\mathbf{0.882}$\\\hline
Concrete & $0.857$ & $\mathbf{0.896}$ & $0.863$ & $\mathbf{0.896}$\\\hline
Wine & $0.423$ & $\mathbf{0.481}$ & $0.416$ & $\mathbf{0.486}$\\\hline
Yacht & $0.989$ & $\mathbf{0.993}$ & $0.988$ & $\mathbf{0.993}$\\\hline
Regression & $0.450$ & $\mathbf{0.455}$ & $0.447$ & $\mathbf{0.464}$\\\hline
Sparse & $0.529$ & $\mathbf{0.641}$ & $0.536$ & $\mathbf{0.663}$\\\hline
\end{tabular}
\label{t:RF_Att_L_7}%
\end{table}%

\section{Concluding remarks}

New attention-based RF models proposed in the paper have supplemented the
class attention models incorporated into machine learning models different
from neural networks \cite{Konstantinov-Utkin-22d,Utkin-Konstantinov-22}.
Moreover, the proposed models do not use gradient-based algorithms to learn
attention parameters, and their training is based on solving the quadratic
optimization problem with linear constraints. This peculiarity significantly
simplifies the training process.

It is interesting to point out that computing the attention weights in leaves
of trees is a very simple task from the computational point of view. At the
same time, this simple modification leads to the crucial improvement of the RF
models. Numerical results with real data have demonstrated this improvement.
This fact motivates us to continue developing attention-based modifications of
machine learning models in different directions. First of all, the same
approach can be applied to the gradient boosting machine \cite{Friedman-2002}.
The first successful attempt to use the attention mechanism in the gradient
boosting machine with decision trees as base learners has been carried out in
\cite{Konstantinov-Utkin-22d}. This attempt has been shown that the boosting
model can be improved by adding the attention component. An idea of the
\textquotedblleft leaf\textquotedblright\ attention and the optimization over
parameters of kernels can be directly transferred to the gradient boosting
machine. This is a direction for further research.

One of the important results presented in the paper is the usage of a specific
mixture of contamination models which can be regarded as a variant of the
well-known multi-head attention \cite{Vaswani-etal-17}, where each
\textquotedblleft head\textquotedblright\ is defined by the kernel parameter.
However, values of the parameter are selected in accordance with a predefined
set. Therefore, the next direction for research is to consider randomized
procedures for selecting values of the parameter.

The proposed models consider only a single leaf of a tree for every example
and implement the \textquotedblleft leaf\textquotedblright\ attention in this
leaf. However, they do not take into account neighbor leaves which also may
provide useful information for improving the models. The corresponding
modification can be also regarded as another direction for research.

It has been shown in \cite{Utkin-Konstantinov-22} that attention-based RFs
allow us to interpret predictions by using the attention weights. The
introduced two-level attention mechanisms may also improve interpretability of
RFs by taking into account additional factors. The corresponding procedures of
the interpretation is also a direction for further research.

Finally, we have developed the proposed modifications by using the Huber's
$\epsilon$-contamination model and the mixture of the models. Another problem
of interest is to consider different available statistical models
\cite{Walley91} and their mixtures. A proper choice of the mixture components
may significantly improve the whole attention-based RF. This is also a
direction for further research.

\section*{Acknowledgement}

The research is partially funded by the Ministry of Science and Higher
Education of the Russian Federation under the strategic academic leadership
program 'Priority 2030' (Agreement N 075-15-2021-1333 dd 30.09.2021).

\bibliographystyle{unsrt}
\bibliography{Attention,Boosting,Classif_bib,Deep_Forest,Explain,Imprbib,MIL,MYBIB,MYUSE}

\begin{thebibliography}{10}

\bibitem{Chaudhari-etal-2019}
S.~Chaudhari, V.~Mithal, G.~Polatkan, and R.~Ramanath.
\newblock An attentive survey of attention models.
\newblock arXiv:1904.02874, Apr 2019.

\bibitem{Correia-Colombini-21a}
A.S. Correia and E.L. Colombini.
\newblock Attention, please! {A} survey of neural attention models in deep
  learning.
\newblock arXiv:2103.16775, Mar 2021.

\bibitem{Correia-Colombini-21}
A.S. Correia and E.L. Colombini.
\newblock Neural attention models in deep learning: Survey and taxonomy.
\newblock arXiv:2112.05909, Dec 2021.

\bibitem{Lin-Wang-etal-21}
T.~Lin, Y.~Wang, X.~Liu, and X.~Qiu.
\newblock A survey of transformers.
\newblock arXiv:2106.04554, Jul 2021.

\bibitem{Niu-Zhong-Yu-21}
Z.~Niu, G.~Zhong, and H.~Yu.
\newblock A review on the attention mechanism of deep learning.
\newblock {\em Neurocomputing}, 452:48--62, 2021.

\bibitem{Breiman-2001}
L.~Breiman.
\newblock Random forests.
\newblock {\em Machine learning}, 45(1):5--32, 2001.

\bibitem{Friedman-2001}
J.H. Friedman.
\newblock Greedy function approximation: A gradient boosting machine.
\newblock {\em Annals of Statistics}, 29:1189--1232, 2001.

\bibitem{Friedman-2002}
J.H. Friedman.
\newblock Stochastic gradient boosting.
\newblock {\em Computational statistics \& data analysis}, 38(4):367--378,
  2002.

\bibitem{Konstantinov-Utkin-22d}
A.V. Konstantinov, L.V. Utkin, and S.R. Kirpichenko.
\newblock {AGB}oost: Attention-based modification of gradient boosting machine.
\newblock In {\em 31st Conference of Open Innovations Association (FRUCT)},
  pages 96--101. IEEE, 2022.

\bibitem{Utkin-Konstantinov-22}
L.V. Utkin and A.V. Konstantinov.
\newblock Attention-based random forest and contamination model.
\newblock {\em Neural Networks}, 2022.
\newblock In press.

\bibitem{Zhang2021dive}
A.~Zhang, Z.C. Lipton, M.~Li, and A.J. Smola.
\newblock Dive into deep learning.
\newblock {\em arXiv:2106.11342}, Jun 2021.

\bibitem{Nadaraya-1964}
E.A. Nadaraya.
\newblock On estimating regression.
\newblock {\em Theory of Probability \& Its Applications}, 9(1):141--142, 1964.

\bibitem{Watson-1964}
G.S. Watson.
\newblock Smooth regression analysis.
\newblock {\em Sankhya: The Indian Journal of Statistics, Series A}, pages
  359--372, 1964.

\bibitem{Huber81}
P.J. Huber.
\newblock {\em Robust Statistics}.
\newblock Wiley, New York, 1981.

\bibitem{Utkin-Konstantinov-22c}
L.V. Utkin and A.V. Konstantinov.
\newblock Attention and self-attention in random forests.
\newblock arXiv:2207.04293, Jul 2022.

\bibitem{Vaswani-etal-17}
A.~Vaswani, N.~Shazeer, N.~Parmar, J.~Uszkoreit, L.~Jones, A.N. Gomez,
  L.~Kaiser, and I.~Polosukhin.
\newblock Attention is all you need.
\newblock In {\em Advances in Neural Information Processing Systems}, pages
  5998--6008, 2017.

\bibitem{Geurts-etal-06}
P.~Geurts, D.~Ernst, and L.~Wehenkel.
\newblock Extremely randomized trees.
\newblock {\em Machine learning}, 63:3--42, 2006.

\bibitem{Liu-Huang-etal-21}
F.~Liu, X.~Huang, Y.~Chen, and J.A. Suykens.
\newblock Random features for kernel approximation: A survey on algorithms,
  theory, and beyond.
\newblock arXiv:2004.11154v5, Jul 2021.

\bibitem{Choromanski-etal-21}
K.~Choromanski, V.~Likhosherstov, D.~Dohan, X.~Song, A.~Gane, T.~Sarlos,
  P.~Hawkins, J.~Davis, A.~Mohiuddin, L.~Kaiser, D.~Belanger, L.~Colwell, and
  A.~Weller.
\newblock Rethinking attention with performers.
\newblock In {\em 2021 International Conference on Learning Representations},
  2021.

\bibitem{Choromanski-etal-21a}
K.~Choromanski, H.~Chen, H.~Lin, Y.~Ma, A.~Sehanobish, D.~Jain, M.S. Ryoo,
  J.~Varley, A.~Zeng, V.~Likhosherstov, D.~Kalachnikov, V.~Sindhwani, and
  A.~Weller.
\newblock Hybrid random features.
\newblock arXiv:2110.04367v2, Oct 2021.

\bibitem{Ma-Kong-etal-21}
X.~Ma, X.~Kong, S.~Wang, C.~Zhou, J.~May, H.~Ma, and L.~Zettlemoyer.
\newblock Luna: Linear unified nested attention.
\newblock arXiv:2106.01540, Nov 2021.

\bibitem{Schlag-etal-2021}
I.~Schlag, K.~Irie, and J.~Schmidhuber.
\newblock Linear transformers are secretly fast weight programmers.
\newblock In {\em International Conference on Machine Learning 2021}, pages
  9355--9366. PMLR, 2021.

\bibitem{Peng-Pappas-etal-21}
H.~Peng, N.~Pappas, D.~Yogatama, R.~Schwartz, N.~Smith, and L.~Kong.
\newblock Random feature attention.
\newblock In {\em International Conference on Learning Representations (ICLR
  2021)}, pages 1--19, 2021.

\bibitem{Brauwers-Frasincar-22}
G.~Brauwers and F.~Frasincar.
\newblock A general survey on attention mechanisms in deep learning.
\newblock arXiv:2203.14263, Mar 2022.

\bibitem{Goncalves-etal-2022}
T.~Goncalves, I.~Rio-Torto, L.F. Teixeira, and J.S. Cardoso.
\newblock A survey on attention mechanisms for medical applications: are we
  moving towards better algorithms?
\newblock arXiv:2204.12406, Apr 2022.

\bibitem{Santana-Colombini-21}
A.~Santana and E.~Colombini.
\newblock Neural attention models in deep learning: Survey and taxonomy.
\newblock arXiv:2112.05909, Dec 2021.

\bibitem{Soydaner-22}
D.~Soydaner.
\newblock Attention mechanism in neural networks: Where it comes and where it
  goes.
\newblock arXiv:2204.13154, Apr 2022.

\bibitem{Xu-Wei-etal-22}
Y.~Xu, H.~Wei, M.~Lin, Y.~Deng, K.~Sheng, M.~Zhang, F.~Tang, W.~Dong, F.~Huang,
  and C.~Xu.
\newblock Transformers in computational visual media: A survey.
\newblock {\em Computational Visual Media}, 8(1):33--62, 2022.

\bibitem{Kim-Kim-Moon-Ahn-2011}
H.~Kim, H.~Kim, H.~Moon, and H.~Ahn.
\newblock A weight-adjusted voting algorithm for ensemble of classifiers.
\newblock {\em Journal of the Korean Statistical Society}, 40(4):437--449,
  2011.

\bibitem{Ronao-Cho-2015}
C.A. Ronao and S.-B. Cho.
\newblock Random forests with weighted voting for anomalous query access
  detection in relational databases.
\newblock In {\em Artificial Intelligence and Soft Computing. ICAISC 2015},
  volume 9120 of {\em Lecture Notes in Computer Science}, pages 36--48, Cham,
  2015. Springer.

\bibitem{Utkin-Konstantinov-etal-20}
L.V. Utkin, A.V. Konstantinov, V.S. Chukanov, and A.A. Meldo.
\newblock A new adaptive weighted deep forest and its modifications.
\newblock {\em International Journal of Information Technology \& Decision
  Making}, 19(4):963--986, 2020.

\bibitem{Utkin-etal-2019}
L.V. Utkin, A.V. Konstantinov, V.S. Chuknov, M.V. Kots, M.A. Ryabinin, and A.A.
  Meldo.
\newblock A weighted random survival forest.
\newblock arXiv:1901.00213, Jan 2019.

\bibitem{Winham-etal-2013}
S.J. Winham, R.R. Freimuth, and J.M. Biernacka.
\newblock A weighted random forests approach to improve predictive performance.
\newblock {\em Statistical Analysis and Data Mining}, 6(6):496--505, 2013.

\bibitem{Xuan-etal-18}
S.~Xuan, G.~Liu, and Z.~Li.
\newblock Refined weighted random forest and its application to credit card
  fraud detection.
\newblock In {\em Computational Data and Social Networks}, pages 343--355,
  Cham, 2018. Springer International Publishing.

\bibitem{Zhang-Wang-21}
X.~Zhang and M.~Wang.
\newblock Weighted random forest algorithm based on bayesian algorithm.
\newblock In {\em Journal of Physics: Conference Series}, volume 1924, pages
  1--6. IOP Publishing, 2021.

\bibitem{Abellan-Moral-2003}
J.~Abellan and S.~Moral.
\newblock Building classification trees using th building classification trees
  using the total uncertainty criterion.
\newblock {\em International Journal of Intelligent Systems},
  18(12):1215--1225, 2003.

\bibitem{Abellan-etal-2017}
J.~Abellan, C.J. Mantas, and J.G. Castellano.
\newblock A random forest approach using imprecise probabilities.
\newblock {\em Knowledge-Based Systems}, 134:72--84, 2017.

\bibitem{Abellan-etal-2018}
J.~Abellan, C.J. Mantas, J.G. Castellano, and S.~Moral-Garcia.
\newblock Increasing diversity in random forest learning algorithm via
  imprecise probabilities.
\newblock {\em Expert Systems With Applications}, 97:228--243, 2018.

\bibitem{Mantas-Abellan-2014}
C.J. Mantas and J.~Abellan.
\newblock Analysis and extension of decision trees based on imprecise
  probabilities: {A}pplication on noisy data.
\newblock {\em Expert Systems with Applications}, 41(5):2514--2525, 2014.

\bibitem{Moral-Garcia-etal-2020}
S.~Moral-Garcia, C.J. Mantas, J.G. Castellano, M.D. Benitez, and J.~Abellan.
\newblock Bagging of credal decision trees for imprecise classification.
\newblock {\em Expert Systems with Applications}, 141(Article 112944):1--9,
  2020.

\bibitem{Utkin-Wiencierz-13}
L.V. Utkin and A.~Wiencierz.
\newblock An imprecise boosting-like approach to regression.
\newblock In Fabio Cozman, Therry Denoeux, S{\'e}bastien Destercke, and Teddy
  Seidfenfeld, editors, {\em ISIPTA '13, Proceedings of the Eighth
  International Symposium on Imprecise Probability: Theories and Applications},
  pages 345--354, Compiegne, France, 2013. SIPTA.

\bibitem{Utkin-Kovalev-Coolen-2020}
L.V. Utkin, M.S. Kovalev, and F.~Coolen.
\newblock Imprecise weighted extensions of random forests for classification
  and regression.
\newblock {\em Applied Soft Computing}, 92(Article 106324):1--14, 2020.

\bibitem{Bahdanau-etal-14}
D.~Bahdanau, K.~Cho, and Y.~Bengio.
\newblock Neural machine translation by jointly learning to align and
  translate.
\newblock arXiv:1409.0473, Sep 2014.

\bibitem{Luong-etal-2015}
T.~Luong, H.~Pham, and C.D. Manning.
\newblock Effective approaches to attention-based neural machine translation.
\newblock In {\em Proceedings of the 2015 Conference on Empirical Methods in
  Natural Language Processing}, pages 1412--1421. The Association for
  Computational Linguistics, 2015.

\bibitem{Dua:2019}
D.~Dua and C.~Graff.
\newblock {UCI} machine learning repository, 2017.

\bibitem{Demsar-2006}
J.~Demsar.
\newblock Statistical comparisons of classifiers over multiple data sets.
\newblock {\em Journal of Machine Learning Research}, 7:1--30, 2006.

\bibitem{Walley91}
P.~Walley.
\newblock {\em Statistical Reasoning with Imprecise Probabilities}.
\newblock Chapman and Hall, London, 1991.

\end{thebibliography}

\end{document}